%% file: main.tex
\documentclass[conference]{IEEEtran}
\IEEEoverridecommandlockouts
\usepackage{cite}
\usepackage{amsmath,amssymb,amsfonts}
\usepackage{algorithmic}
\usepackage{graphicx}
\usepackage{textcomp}
\usepackage{xcolor}
\usepackage{latexsym}
\usepackage{amssymb}
\usepackage{amsmath}
\usepackage{amsthm}
\usepackage{booktabs}
\usepackage{enumitem}
\usepackage{graphicx}
\usepackage{color}
\usepackage{multirow}
\usepackage{xcolor}
\usepackage{tabularx}
\usepackage[para,online]{threeparttable}

\def\BibTeX{{\rm B\kern-.05em{\sc i\kern-.025em b}\kern-.08em
    T\kern-.1667em\lower.7ex\hbox{E}\kern-.125emX}}
\begin{document}

\title{CrowdVLM-R1: Expanding R1 Ability to Vision
Language Model for Crowd Counting using Fuzzy Group
Relative Policy Reward\\
}

\author{\IEEEauthorblockN{1\textsuperscript{st} Zhiqiang Wang}
\IEEEauthorblockA{\textit{Florida Atlantic University}\\
Boca Raton, USA \\
zwang2022@fau.edu}
\and
\IEEEauthorblockN{2\textsuperscript{nd} Pengbin Feng}
\IEEEauthorblockA{\textit{University of Southern California}\\
Los Angeles, USA \\
fengpengbin.apply@gmail.com}
\and
\IEEEauthorblockN{3\textsuperscript{rd} Yanbin Lin}
\IEEEauthorblockA{\textit{Florida Atlantic University}\\
Boca Raton, USA \\
liny2020@fau.edu}
\and
\IEEEauthorblockN{4\textsuperscript{th} Shuzhang Cai}
\IEEEauthorblockA{\textit{University of Texas at Dallas}\\
Texas, USA \\
caishuzhang@gmail.com}
\and
\IEEEauthorblockN{5\textsuperscript{th} Zongao Bian}
\IEEEauthorblockA{\textit{Georgia Institute of Technology}\\
Atlanta, USA \\
zongao.bian@gatech.edu}
\and
\IEEEauthorblockN{6\textsuperscript{th} Jinghua Yan}
\IEEEauthorblockA{\textit{University of Utah}\\
Salt Lake City, USA \\
jhyan@cs.utah.edu}
\and
\IEEEauthorblockN{7\textsuperscript{th} Xingquan Zhu}
\IEEEauthorblockA{\textit{Florida Atlantic University}\\
Boca Raton, USA \\
xzhu3@fau.edu}
}

\maketitle

\begin{abstract}
We propose CrowdVLM-R1, which expands the R1 base model for accurate crowd counting, using a novel framework that integrates the fuzzy group relative policy optimization reward function (FGRPR) to enhance learning efficiency. Unlike the conventional binary (0/1) accuracy reward, our fuzzy reward model, FGRPR, which contains both format and precision rewards, provides nuanced incentives to encourage the R1 model to learn to adjust policies towards precise outputs. Supervised fine-tuning (SFT) is also integrated for the CrowdVLM-R1 model to learn from a handful of inputs to enable both in-domain and out-of-domain counting. Experimental results demonstrate that GRPO with a standard binary accuracy reward underperforms compared to SFT. In contrast, FGRPR, applied to Qwen2.5-VL-(3B/7B), surpasses all baseline models, including GPT-4o, LLaMA2-70B and SFT, in five domain datasets. For out-of-domain datasets, FGRPR achieves performance comparable to SFT but excels when target values are larger, as its fuzzy reward function assigns higher rewards to closer approximations. This approach is broadly applicable to tasks where the precision of the answer is critical. The code and data are available at: https://github.com/yeyimilk/CrowdVLM-R1
\end{abstract}

\input{sec/method_framework}

\input{sec/introduction}

\input{sec/related}

\input{sec/method}

\input{sec/experiement}
\input{sec/limitation}

\input{sec/conclusion}

\section*{Acknowledgment}
This work has been supported in part by the US National Science Foundation (NSF) under Grant No. IOS-2430224.

\bibliographystyle{IEEEtran}
\bibliography{reference}

\end{document}

%% file: sec/method_framework.tex
\begin{figure*}
    \centering
    \includegraphics[width=1\linewidth]{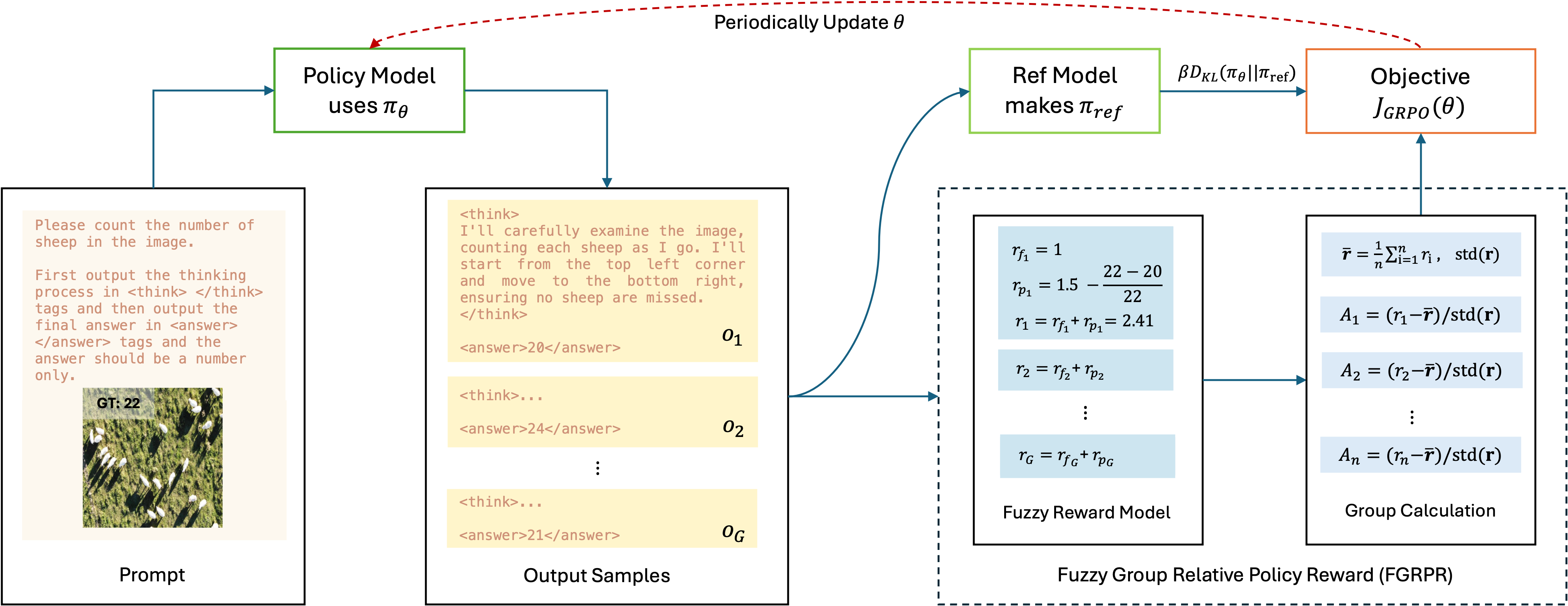}
    \caption{An overview of our framework, which integrates group relative policy optimization (GRPO) and the proposed fuzzy group relative policy reward (FGRPR) to train a visual language model (VLM) for crowd counting tasks. From left to right, an image and prompt are input to a base LLM to obtain initial counting outcomes (The ground truth value on the top left corner of the input image is just an indicator number for visualization purposes. It is only used for rewards calculation, but will not send to LLM as part of the input. The instruction model uses a policy model, parameterized by $\theta$, to obtain policy and generate outputs. The proposed fuzzy group relative policy reward is then used to adjust the reward for the individual image. The adjusted reward is used to calculate the objective function value, which in turn updates the $\theta$ parameters for supervised fine-tuning.   
    }
    \label{fig: framework}
\end{figure*}

%% file: sec/introduction.tex
\section{Introduction}

Recently, DeepSeek R1 \cite{guo2025deepseek} has drawn much attention among advances in large language models (LLMs), as it demonstrates how reinforcement learning (RL) can be the primary driver of reasoning. The model is a Mixture-of-Experts (MoE) transformer trained using the GRPO algorithm \cite{shao2024deepseekmath}; specifically, it applies simple rule-based rewards to guide the model toward complex reasoning skills.

While DeepSeek R1 is an LLM, its training paradigm can be extended to vision-language models (VLMs). There are various types of visual tasks, such as crowd counting \cite{zhang2015cross,liu2019context,liang2023crowdclip}, image classification \cite{li2014medical,menon2022visual,qi2024benchmarking}, and visual question answering \cite{antol2015vqa,guo2023images}; in the context of VLM, they require the model to go beyond simple pattern recognition and perform precise reasoning. Latest works explore applying RL for visual tasks, one of which is VLM-R1 \cite{shen2025vlmr1}, which assigns rewards based on the intersection over union (IoU) score - a binary reward of 1 if IoU $>$ 0.5, and 0 otherwise. However, this approach does not distinguish between near-correct and marginally correct detections (e.g., an IoU of 0.55 vs. 0.95), then better results do not receive higher rewards. This suggests the need for more fine-grained reward schemes when applying RL to vision-based reasoning tasks.

In this paper, we develop a novel approach based on GRPO to handle crowd counting tasks. It has been a fundamental yet highly challenging problem and the primary goal is to accurately estimate the number of instances of a particular object. In such scenarios, objects can overlap or vary in size, and they usually fade into complex backgrounds. The VLM must avoid either double-counting or missing partially covered objects. This demands strong spatial reasoning and fine-grained attention.

Therefore, we explore key research questions regarding the application of RL to VLMs for the counting task:

\begin{itemize}
    \item \textbf{RQ1:} How effective is GRPO for VLMs for crowd counting tasks?

    \item \textbf{RQ2:} How can a better reward function be designed to enhance the numerical reasoning ability of VLMs?
    
    \item \textbf{RQ3:} How to improve VLMs for both within- and cross-domain counting tasks?
\end{itemize}

To address RQ1, we first conduct a systematic investigation into the impact of RL on VLMs. Specifically, we adapt GRPO, originally used in DeepSeek R1, for fine-tuning mainstream VLMs. Our experiments confirm that GRPO is highly feasible for the counting task in VLMs. However, it requires a customized reward function, as conventional reward function has certain limitations. 

Issues like this highlight the need to address our RQ2. A well-structured reward function should provide feedback with finer granularity and ensure that slight improvements in counting accuracy are properly incentivized rather than treated as binary outcomes. We solve this research question by introducing a fuzzy group relative policy reward (FGRPR). Our reward function considers both the model's output format and the numerical precision of its predictions.  The format reward is designed as a binary signal to encourage the model to provide an actual count for the scene. Meanwhile, the accuracy reward is based on the difference between the predicted count and the ground truth, guiding the model toward more precise predictions by penalizing deviations from the correct count.

To address RQ3, we create a new, comprehensive counting dataset by selectively combining multiple existing image datasets. Given the complexity of the counting task, which involves a wide variety of scenes such as sparse and dense object distributions, as well as large and small objects, we ensure that our dataset captures this diversity. Our goal is to enable the model to adapt to various scene types. Additionally, to evaluate the model's out-of-domain capabilities, we incorporate the Manatee dataset, which will be used exclusively for testing to assess the model’s zero-shot counting accuracy.

The contributions of this work are as follows:

\begin{itemize}
    \item We adapt reinforcement learning (GRPO) to train VLMs for crowd counting and propose CrowdVLM-R1, which achieves state-of-the-art performance on the counting task;
    \item We design a novel fuzzy reward function that takes into account both the format and counting precision, enhancing the model's ability to provide accurate predictions;
    \item We introduce a new comprehensive counting dataset, offering diverse scenes to assess the model's counting capabilities for in-domain and out-of-domain scenes.
\end{itemize}

%% file: sec/related.tex
\section{Related Work}

\noindent{\bf Traditional methods for Crowd Counting.} Early research on Crowd Counting can be grouped into three categories: detection-based, regression-based, and density estimation-based methods. Detection-based approaches typically utilize detectors for a specific object, e.g., human head or animal body, and perform counting based on the detection results \cite{wang2011automatic,wu2005detection}. These methods have achieved good accuracy for sparse scenes, but struggle under dense or occluded scenarios. Regression-based methods, on the other hand, map the image patches directly to the count of objects via regression techniques \cite{chen2013cumulative}. Density estimation methods get rid of learning to localize individual objects by estimating the count with density maps \cite{wang2012crowd, fiaschi2012learning}.

With the success of Convolution Neural Networks (CNNs) in multiple computer vision tasks, researchers have started to leverage CNNs to learning non-linear mappings from crowd images to counts. CNNs have been proven to extract features of higher quality than traditional methods for many vision tasks, including counting \cite{sermanet2013overfeat}. Among numerous CNN-based approaches, early works such as \cite{wang2015deep, fu2015fast} use CNN models to perform end-to-end regression without significant modification of the model architecture. Subsequent research has focused on enhancing model architectures \cite{babu2017switching, sindagi2017cnn,sermanet2012convolutional, kumagai2017mixture}. Examples include multi-stage CNNs \cite{sermanet2012convolutional}, which are designed to adapt to counting scenes with different scales and distortion, and Mixture of CNNs \cite{kumagai2017mixture} with a mixture-of-exptert approach where each CNNs is specialized to a different scene. 

\noindent{\bf Vision-Language Models (VLMs) for Counting.} The need for supervision and labeling of datasets limits the generalizability of both CNN-based and non-CNN-based methods. Recent vision-language pretraining works \cite{radford2021learning}\cite{liu2024grounding} do not rely on supervised training by learning from an extremely large number of image-text pairs. VLMs generates high-quality visual-textual features, which are adaptable to a broad spectrum of open-world downstream tasks. Recent VLM-based counting methods have demonstrated superior performance compared to class-specific approaches in counting accuracy \cite{liu2022countr, you2023few, kang2024vlcounter, djukic2023low, dai2024referring, amini2023open}. These approaches establish a joint embedding space that connects images with textual inputs, and hence enable models to comprehend general concepts learned during extensive pretraining. 

\noindent{\bf Reasoning enhancement for Large Language Models.}
As LLMs' capabilities advance, there is a growing research focus on their reasoning abilities. A recent work, DeepSeek R1 \cite{guo2025deepseek}, utilizes pure reinforcement learning with rule-based reward and demonstrates strong performance on a series of reasoning tasks. This approach can be naturally extended to the multimodality setting and applied to VLMs \cite{liu2025visual,pan2025medvlm,zhan2025vision,zhang2025r1, xu2025survey}.

%% file: sec/method.tex
\section{method}
\begin{figure*}
    \includegraphics[width=1\linewidth]{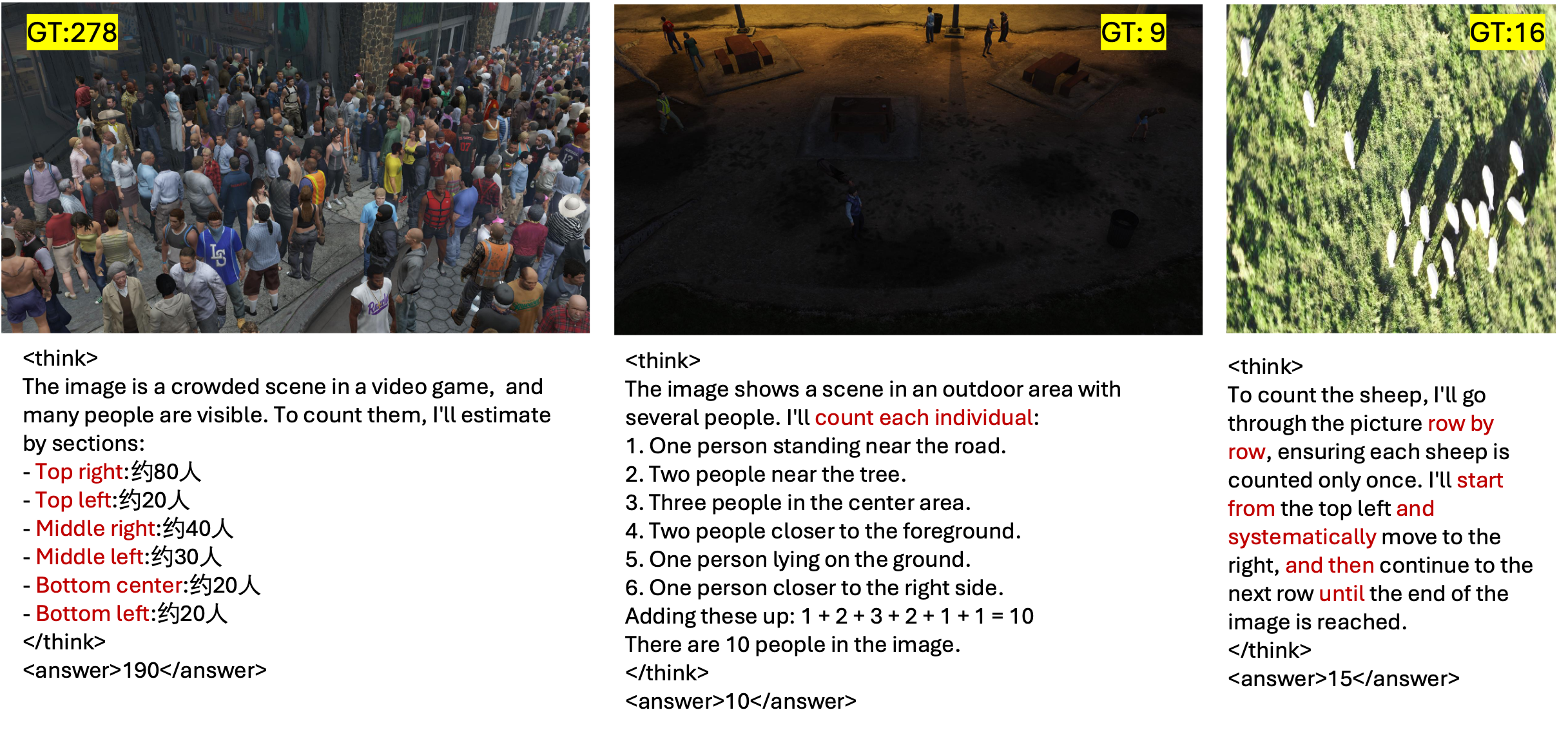}
    \caption{Examples of model outputs during the training stage, showing various counting strategies. In the left image, where the crowd is dense and difficult to count, the model segments the scene into six regions and estimates the number of individuals in each. In the middle image, with only a few people, the model counts them individually. In the right image, which has a moderate density of sheep, the model scans row by row to ensure each entity is counted once. These strategies resemble human approaches to counting in different scenarios.}
    \label{fig:process_sample_o}
\end{figure*}

In this section, we present our approach to improving VLM-based counting through reinforcement learning. We begin by introducing Group Relative Policy Optimization (GRPO) and highlighting its advantages over previous RL methods. Next, we propose a novel fuzzy group relative policy reward (FGRPR), which refines the reward structure by accounting for both output format adherence and numerical precision. Finally, we outline our overall framework, which integrates GRPO and FGRPR into the training pipeline of our proposed model, CrowdVLM-R1, to improve its counting performance across diverse datasets and real-world scenarios.

\subsection{Group Relative Policy Optimization}
GRPO \cite{shao2024deepseekmath} is an RL algorithm derived as a variant of Proximal Policy Optimization (PPO) \cite{schulman2017proximal}. This algorithm was introduced by the Deepseek group, extending the PPO framework to incorporate group-based optimization strategies. The value function used in PPO is typically implemented as a separate model of comparable size to the policy model, introducing significant memory and computational overhead. By utilizing the average reward of multiple sampled outputs generated in response to the same input as the baseline, GRPO eliminates the need for an additional value function approximation. This approach reduces both computational complexity and memory overhead, while maintaining effective variance reduction in policy gradient updates.

For standardized GRPO, given an input question $q$, a group of $G$ outputs $\mathbf{o}=\{o_1, o_2, \dots, o_G \} $ are generated from a sampling process determined by the old policy $\pi_{\theta_{\text{old}}}$, parameterized by $\theta$, to answer the question $q$. Let $\textbf{r} = \{ r_1, r_2, \dots, r_G \}$ denote $G$ rewards in the group and $\bar{\textbf{r}}$ be the group average reward, the advantages $\hat{A}_{i,t}$ of all tokens in the output as the normalized reward will be 
\begin{equation}
\hat{A}_{i,t} = \tilde{r}_i = \frac{r_i - \bar{\textbf{r}}}{\text{std}(\mathbf{r})}
\end{equation}

The policy model will be optimized by maximizing the objective:
\begin{equation}
\begin{aligned}
 \mathcal{J}_{GRPO}(\theta) &= \mathbb{E}_{ q \sim P(Q), \{ o_i \}_{i=1}^{G} \sim \pi_{\theta_{\text{old}}}(O | q) }   \\
&  \frac{1}{G} \sum_{i=1}^{G} \frac{1}{|o_i|} \sum_{t=1}^{|o_i|}\{ \min [\frac{\pi_{\theta}(o_{i,t} | q, o_{i,<t})}{\pi_{\theta_{\text{old}}}(o_{i,t} | q, o_{i,<t})} \hat{A}_{i,t}, \\
& \text{clip} \left( \frac{\pi_{\theta}(o_{i,t} | q, o_{i,<t})}{\pi_{\theta_{\text{old}}}(o_{i,t} | q, o_{i,<t})}, 1 - \epsilon, 1 + \epsilon \right) \hat{A}_{i,t}] \\ 
&- \beta D_{KL} \left[ \pi_{\theta} || \pi_{\text{ref}} \right] \}\\
\end{aligned}
\label{j_theta}
\end{equation}
where $\epsilon$ and $\beta$ are hyperparameters, and $\hat{A}_{i,t}$ represents the advantage computed based on the relative 
rewards of outputs within each group. $\pi_{\theta}$ is the current policy to be optimized, generated by the trainable policy model, typically an LLM or sequence model, parameterized by $\theta$. $\pi_{old}$ is the behavior policy used for sampling, which is obtained from a frozen model of the policy $\pi_{\theta}$ trained in a previous iteration. $\pi_{ref}$ is the reference policy for the KL penalty, which is typically from the SFT model (also a LLM or sequence model).

GRPO determines which strategies should be retained or improved by comparing the performance of a group of samples, rather than relying on the absolute reward of a single sample. This approach is analogous to the process of natural selection, where only strategies that outperform the average are allowed to survive and evolve further. This group-based method makes GRPO more stable when handling diverse tasks, as it no longer depends on the performance of a single sample but instead leverages the collective intelligence of the group to evaluate the quality of strategies.

PPO employs a clipped objective function to limit the magnitude of policy updates. However, this clipping mechanism can sometimes overly restrict updates, leading to slower convergence. On the other hand, Trust Region Policy Optimization (TRPO) ensures the stability of policy updates through a strict KL divergence constraint, but its optimization process requires second-order methods (e.g., conjugate gradient), resulting in high computational costs that make it unsuitable for large-scale reinforcement learning tasks. GRPO addresses these limitations by combining clipping mechanisms with a KL divergence term, ensuring a balance between exploration and stability. This balance makes GRPO more reliable in complex tasks, as it maintains both stability and consistency during optimization. While PPO's fixed clipping range may limit the convergence speed of policies, GRPO introduces a dynamic adjustment mechanism (e.g., through a beta parameter) to adaptively control the policy update range, effectively adjusting the KL divergence to achieve stable and efficient policy updates.

In previous work like VLM-R1 \cite{shen2025vlmr1}, a typical GRPO accuracy reward is defined as

\begin{equation}
    r_{a} = \begin{cases}
1, & \frac{\left|\hat{y} - y \right|}{y} < 0.5 \\
0, & \text{otherwise }
\end{cases}
\label{equ: ra}
\end{equation}

where $\hat{y}$ is the count predicted by the model and $y$ denotes the real count. This is a "fixed reward" that assigns 1 to any prediction values lie in the range $(\frac{y}{2}, \frac{3y}{2})$. For example, if the image contains $500$ persons, the prediction values of $260$, $500$, and $740$ share the same reward and have no difference from a model perspective. However, these values are supposed to have different rewards for their different accuracies.

\subsection{FGRPR: Fuzzy Group Relative Policy Reward}
\label{rm}

To enable SFT by using multiple sources of crowd counting data, we introduce the FGRPR method, which employs a fuzzy reward function to gauge the fine-tuning of the R1 model by taking quality of counting, compared to the benchmark, for consideration. The general reward function is defined as
\begin{itemize}
 \item \textbf{Multi-factor Reward:} This reward can describe the task in meaningful dimensions (e.g., format, accuracy, precision, coherence, diversity).
    \item \textbf{Modular Design:} The total reward is a weighted sum:
    \[
       r(s,a) = \sum_{i} \mu_{i}(s, a) r_i(s,a),
    \]
    where each \( r_i(s,a) \) evaluates a specific aspect.
\end{itemize}
For the crowd counting task (specifically in 2 dimensional images), one factor is format adherence, and another is counting precision. Overall, the reward function consists of two parts: \textit{format reward}, $r_f$, and \textit{precision reward}, $r_p$, which are linearly combined in Eq.~(\ref{equ: r}).

\begin{equation}
    r = r_f + r_p
    \label{equ: r}
\end{equation}

\begin{equation}
r(s, a) = \mu_{f}(s, a) \cdot r_{f}(s,a) + \mu_{p}(s, a) \cdot r_{p}(s,a)
\end{equation}

The format reward is straightforward: if the model output format ($\hat{y}_p$) follows the defined format ($\mathbb{F}$), $r_f$ is $1$, otherwise, $r_f$ is $0$. The membership degree $\mu_f(s, a)$ quantifies how closely the output $a$ matches the intended format in state $s$, forming the fuzzy set "format-compliant" (e.g., if the output closely matches but still deviates, $\mu_f$ may be 0.8 rather than a crisp 1). Likewise, $\mu_p(s, a)$ represents the fuzzy set "precision-adequate" (e.g., $\mu_p$ = 0.6 if the numeric answer is close but not exact). For all ablation experiments presented, we set both $\mu_f$ and $\mu_p$ defaults to 1 to equally balance format and precision.

\begin{equation}
    r_f = \begin{cases}
        1, & \text{if } \hat{y}_p = \mathbb{F}\\
        0, & \text{otherwise} 
    \end{cases}
    \label{equ: rf}
\end{equation}

Previous work indicated that the answers can be simply categorized as "Yes" or "No", which means the precision reward can be easily used with binary value—1 or 0. While the crowd counting task also has a ground truth value, we prefer to have a fuzzy reward to represent the model estimated value that it should have a higher reward when the answer is closer to number of objects and lower reward or even 0 if the predicted value is far away from the ground truth value. Thus, the precision reward in our task is defined as shown below.

\begin{equation}
    r_p = \begin{cases}
1.5 - \frac{\left|\hat{y} - y \right|}{y}, & \frac{\left|\hat{y} - y \right|}{y} < 0.5 \\
0, & \text{otherwise }
\end{cases}
\label{equ: rp}
\end{equation}

where $\hat{y}$ is the count predicted by the model and $y$ denotes the real count. The reward values fall within the range $0 \cup (1, 1.5]$. The reason of the range $(1, 1.5]$ instead of $(0.5, 1]$ is motivated by the nature of the counting task, where prediction accuracy is prioritized over formatting. As a result, higher rewards are assigned when the model produces near-correct predictions, emphasizing the importance of accuracy.

\subsection{Framework}

\begin{figure}[h]
    \centering
    \includegraphics[width=0.95\linewidth]{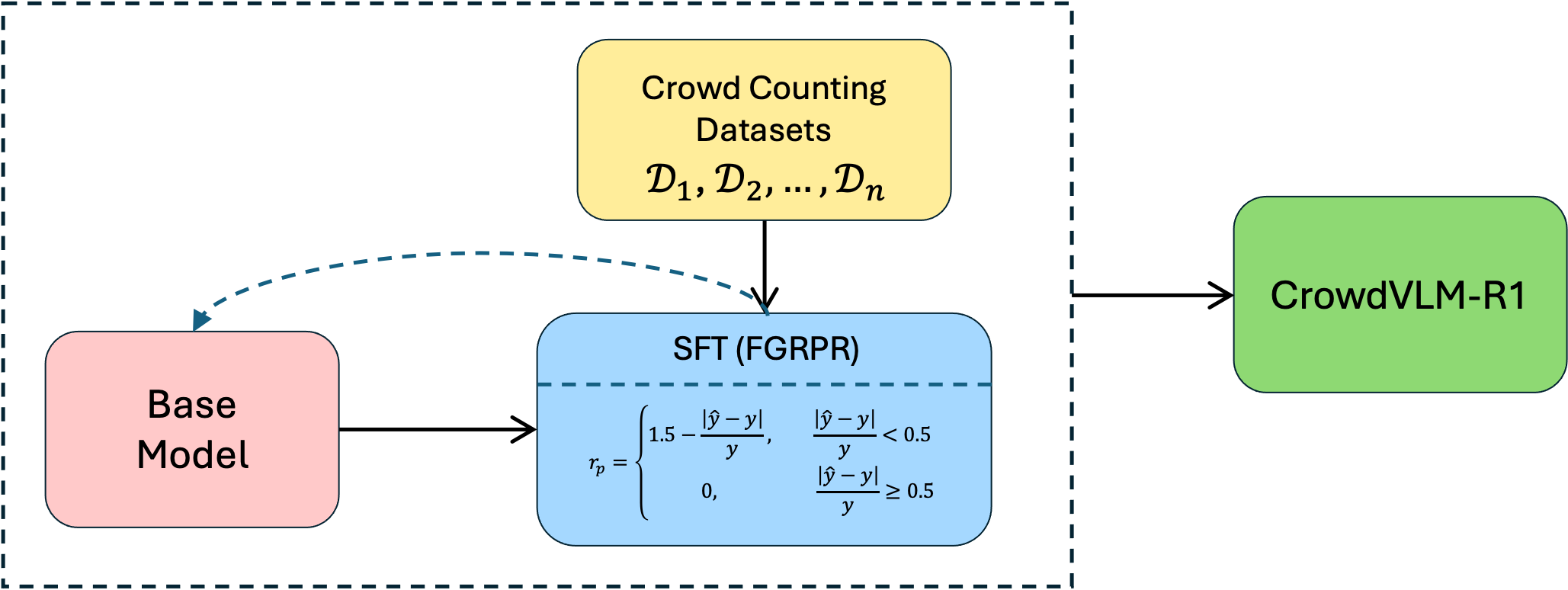}
    \caption{A conceptual view of R1 model expansion to vision
language model for crowd counting (CrowdVLM+R1) using fuzzy group relative policy reward. Multiple crowd-sourced inputs, with ground-truth counting numbers, are used for supervised fine-tuning. The inputs are fed into a base model to generate outcomes, followed by the proposed fuzzy group relative policy reward (FGRPR) to determine individual rewards and system objective function value $\mathcal{J}_{GRPO}(\theta)$. Gradient propagation process is then applied to update $\theta$, and in terms of improving the base model for better counting results.}    
    \label{fig: flow}
\end{figure}

\input{sec/datasets_tab_fig}

Fig. \ref{fig: flow} presents the general idea of training a CrowdVLM-R1 model. Instead of using a labeled dataset to fine-tune the base instruction model to optimize its performance directly, in our method, GRPO is employed during the SFT stage with verifiable rewards (equation \ref{equ: r}), including format reward (equation \ref{equ: rf}) and counting precision reward (equation \ref{equ: rp}).

For details, our proposed framework is shown in Fig. \ref{fig: framework}. A structured prompt comprises a natural language description (including the counting task and the expected output format) and an input image. This prompt is the input into a policy model—implemented as a LLM—which defines the learner policy \( \pi_\theta \) and autoregressively generates a set of candidate outputs \( \{o_1, o_2, \ldots, o_G\} \). Each output \( o_i \) consists of a step-by-step reasoning process and a predicted count corresponding to the target objects in the input image. Subsequently, a reward model evaluates these generated responses, which assigns a scalar reward \( r_i \) to each output. The reward \( r_i \) is composed of two components: a format reward \( r_{f_i} \) and a prediction reward \( r_{p_i} \), as detailed in Section~\ref{rm}. These individual rewards are then aggregated to compute group-wise advantages \( A_i \) for all $G$ outputs, reflecting the relative quality of each sampled response. Our objective function \( \mathcal{J}_{\mathrm{GRPO}}(\theta) \) is formulated using the computed advantages \( A_i \) (for \( i = 1, \ldots, G \)) and a KL-divergence regularization term \( D_{\mathrm{KL}}(\pi_{\theta} \| \pi_{\mathrm{ref}}) \), scaled by a hyperparameter \( \beta \) in Equation~\eqref{j_theta}. Here, in our case, the reference model is an SFT LLM trained with human data. We will minimize the objective/loss function using gradient ascent (or descent on the negative objective). After the calculation, the gradient signal will flow from the objective back into the policy model to update its parameters $\theta$ based on both the relative advantages and the regularization signal. After sufficient interactions, we can finally obtain a model that can accurately count the number of specified targets in a given image.

%% file: sec/datasets_tab_fig.tex
\begin{figure*}[th]
    \centering
    \includegraphics[width=0.9\linewidth]{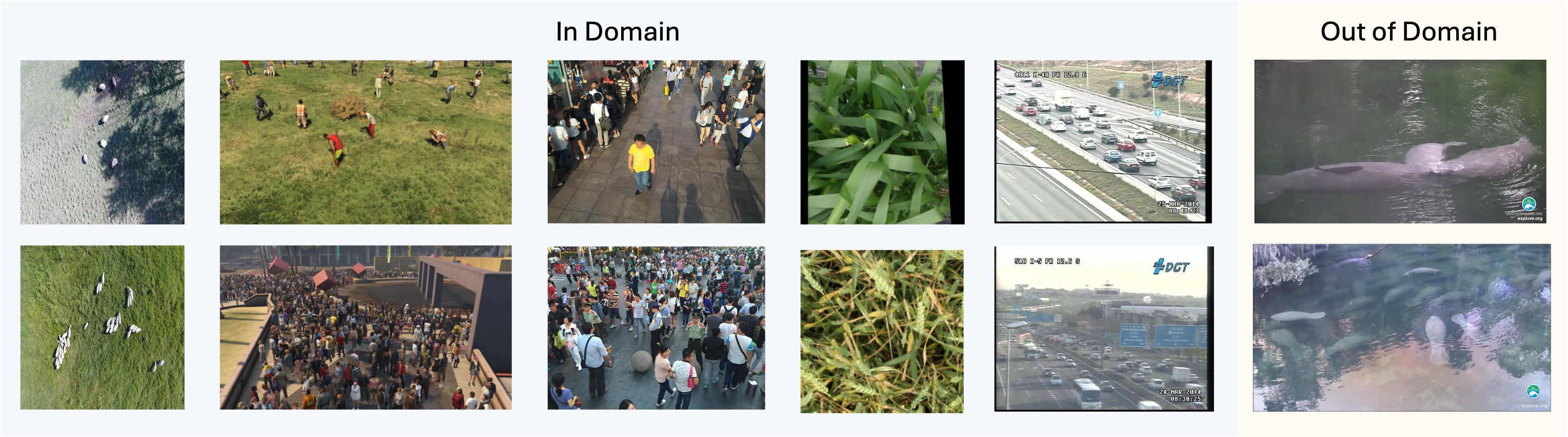}
    \caption{This figure presents the image examples from the test dataset where each column comes from the same original dataset and the first row shows an easier counting task while the second one is more complex within their original data sources.}
    \label{fig:image_example}
\end{figure*}

\begin{table*}[th]
    \centering
        \caption{Dataset Summary}
        \label{tab:datasets}
        \renewcommand{\arraystretch}{1.5}
        \begin{tabular}{c c c c c c c c}
            \hline
             Dataset & Type & Sheep & Video Game Characters & Street Pedestrians & Wheat Heads & Cars & Manatees (out-of-domain) \\
            \hline
            \multirow{4}{*}{Training} & Samples & 1000 & 1000 & 400 & 1000 & 403 & - \\
            \cline{3-8}
            & Min count & 1 & 5 & 12 & 1 & 9 & - \\
            \cline{3-8}
            & Max count & 105 & 3546 & 578 & 125 & 95 & - \\
            \cline{3-8}
            & Mean count & 31 & 326 & 123 & 45 & 34 & - \\
            \hline
            \multirow{4}{*}{Testing} & Samples & 100 & 100 & 100 & 100 & 100 & 100 \\
            \cline{3-8}
             & Min count & 1 & 13 & 9 & 1 & 15 & 1 \\
             \cline{3-8}
             & Max count & 103 & 1709 & 514 & 115 & 107 & 50 \\
             \cline{3-8}
             & Mean count & 29 & 365 & 514 & 49 & 42 & 16 \\
            \hline
        \end{tabular}
    
\end{table*}

%% file: sec/experiement.tex
\section{experiment}
\begin{table*}[h]
    \centering
    \begin{threeparttable}
    \caption{Performance comparison of models across domains.}
    \label{tab:accuracy}
    \renewcommand{\arraystretch}{1.5}
     \begin{tabular}{l p{0.6cm} p{0.7cm} p{0.6cm} p{0.7cm} p{0.6cm} p{0.7cm} p{0.6cm} p{0.7cm} p{0.6cm} p{0.7cm} p{0.6cm} p{0.7cm} p{0.6cm} p{0.7cm}}
        \toprule
        \multirow{4}{*}{Model} & \multicolumn{10}{c}{In-Domain} & \multicolumn{2}{c}{Out-of-Domain} & \multicolumn{2}{c}{\multirow{3}{*}{Overall}} \\
        \cmidrule(lr){2-11} \cmidrule(lr){12-13}
        & \multicolumn{2}{c}{Sheep} & \multicolumn{2}{c}{Characters} & \multicolumn{2}{c}{Pedestrians} & \multicolumn{2}{c}{Wheat Heads} & \multicolumn{2}{c}{Cars} & \multicolumn{2}{c}{Manatees}  \\
        \cmidrule(lr){2-3} \cmidrule(lr){4-5} \cmidrule(lr){6-7} \cmidrule(lr){8-9} \cmidrule(lr){10-11} \cmidrule(lr){12-13} \cmidrule(lr){14-15}
        & MAE & RMSE & MAE & RMSE & MAE & RMSE & MAE & RMSE & MAE & RMSE & MAE & RMSE & MAE & RMSE \\
        \midrule
        Sonnet-3.5 & 8.79 & 13.48 &191.10 & 298.53 &55.96 & 86.77 &36.99 & 44.04 &14.40 & 17.99 &9.44 & 15.12 &52.78 & 128.66 \\
        In-VL2-26B & 14.13 & 22.75 &271.96 & 422.56 &64.19 & 101.23 &33.64 & 41.24 &22.40 & 31.74 &8.22 & 12.99 &69.09 & 178.98 \\
        GPT-4o & 7.12 & 11.73 &243.41 & 504.93 &58.36 & 85.27 &39.79 & 46.58 &17.22 & 23.56 & \colorbox{green!50}{5.82} & \colorbox{green!50}{10.20} &61.95 & 210.24 \\
        LLaMA-70B & \colorbox{green!20}{3.87} & 7.54 &182.40 & 298.53 &28.60 & 38.02 &\colorbox{green!20} {18.18} & \colorbox{green!20} {21.91} &19.24 & 27.31 & \colorbox{green!20}{5.70} & \colorbox{green!20}{10.54} &44.68 & 127.73 \\
        Q-3B & 14.11 & 21.60 &277.67 & 450.83 &76.04 & 122.38 &40.34 & 47.07 &22.83 & 29.71 &11.03 & 16.98 &73.67 & 192.39 \\
        Q-7B & 12.03 & 20.42 &347.44 & 536.23 &121.99 & 166.07 &47.82 & 55.31 &40.24 & 44.24 &11.49 & 17.26 &96.83 & 231.25 \\ \hline
        Q-3B-SFT & 4.19 & \colorbox{green!20}{6.93} &104.34 & 199.79 &33.38 & 52.43 &18.57 & 23.67 &10.68 & \colorbox{green!20}{13.25} &8.15 & 13.40 &29.89 & 85.27 \\

        Q-7B-SFT & 4.86 & 8.24 &97.41 & 224.39 &24.78 & 42.39 & 18.51 & 25.22 &\colorbox{green!20}{11.63} & 14.55 &7.93 & 13.21 &27.52 & 94.20 \\ \hline
        Q-3B-R1 & 5.75 & 9.15 &97.43 & 184.97 &41.74 & 62.71 &25.99 & 30.34 &14.78 & 17.73 &8.92 & 14.39 &32.44 & 81.32 \\
        Q-7B-R1 & 8.04 & 13.20 &145.55 & 261.81 &32.29 & 56.10 &22.68 & 27.40 &14.42 & 18.25 &9.88 & 15.26 &38.81 & 110.44 
        \\ \hline
        Q-3B-FR  & 4.19 & 8.07 & \colorbox{green!50}{85.52} & \colorbox{green!20}{171.28} & \colorbox{green!20}{24.02} & \colorbox{green!20}{38.76} & 18.86 & 23.11 &10.55 & 14.00 &7.81 & 13.25 &25.16 & 72.81 \\
        Q-7B-FR & \colorbox{green!50}{3.33} & \colorbox{green!50}{6.24} & \colorbox{green!20}{88.10} & \colorbox{green!50}{170.17} & \colorbox{green!50}{20.60} & \colorbox{green!50}{40.89} & \colorbox{green!50}{14.52} & \colorbox{green!50}{18.74} & \colorbox{green!50}{9.71} & \colorbox{green!50}{12.47} & 8.16 & 13.23 & \colorbox{green!50}{24.07} & \colorbox{green!50}{72.28} \\
        \bottomrule
    \end{tabular}
    \end{threeparttable}
    \begin{tablenotes}
        \item \hspace{5mm} Q: Qwen-VL-2.5-Instruct; R1: model was trained with GRPO; FR: model was trained with FGRPR
    \end{tablenotes}
\end{table*}

\subsection{Dataset}

The dataset used in this article is derived from existing datasets for counting tasks. The training dataset includes aerial sheep \cite{aerial-sheep_dataset}, video game characters from GTAV\_Head \cite{zhong2024mask}, street pedestrians \cite{zhang2016single}, wheat heads from the Global Wheat Head Detection (GWHD) dataset \cite{david2020global}, and cars from TRANCOS \cite{TRANCOSdataset_IbPRIA2015}. The testing dataset includes an additional dataset, manatee \cite{wang2023counting}, for out-of-domain evaluation.

Fig. \ref{fig:image_example} shows two examples from each original dataset. The images in the first row are clearer, with lower-density objects, making the counting task easier, while the images in the second row depict more complex scenarios with a higher density of objects, increasing task difficulty.

Within our dataset, not all the images from the original datasets are used. If original dataset has train and test sets, then the original train and test samples are picked randomly to use as part of our own train and test sets, respectively. If the original dataset is not split into train and test sets, we first divide it into two parts, and then select images from these subsets to form our train and test sets. In our dataset, we use up to 1000 training images and 100 testing images.

Table \ref{tab:datasets} provides a summary of the datasets used in our study, highlighting the diversity and complexity of the counting task. The dataset encompasses a wide range of scenarios, from relatively simple images containing only a few objects to highly complex scenes with thousands of densely packed, overlapping instances. The variation in object counts, as indicated by the minimum, maximum, and mean values, demonstrates the challenge of accurately estimating quantities across different contexts. Some datasets, such as the video game characters and street pedestrians, include images where objects are densely clustered, making separation and identification difficult. Additionally, environmental noise, including buildings, trees, and other background elements, further complicates detection in real-world scenes. The dataset complexity indicates the challenges of performing a counting task with a single model.
\subsection{Experiment Setup}

\begin{table*}[h]
    \centering
    \begin{threeparttable}
    \caption{Performance comparison of models across ranges.}
    \label{tab:by_range}
    \renewcommand{\arraystretch}{1.5}
    \begin{tabular}{l cc cc cc cc cc}
        \toprule
        \multirow{3}{*}{Model} & \multicolumn{6}{c}{In-Domain} & \multicolumn{4}{c}{Out-Domain} \\
        \cmidrule(lr){2-7} \cmidrule(lr){8-11}
        & \multicolumn{2}{c}{[1-20]} & \multicolumn{2}{c}{[21-50]} & \multicolumn{2}{c}{[51, -]} & \multicolumn{2}{c}{[1-20]} & \multicolumn{2}{c}{[21-50]} \\
        \cmidrule(lr){2-3} \cmidrule(lr){4-5} \cmidrule(lr){6-7} \cmidrule(lr){8-9} \cmidrule(lr){10-11}
        & MAE & RMSE & MAE & RMSE & MAE & RMSE & MAE & RMSE & MAE & RMSE \\
        \midrule
        T-Rex & - & - & - & - & - & - & 4.37 & 6.00 & 32.12 & 33.15 \\
        \hline
        
        Q-3B-SFT & 2.99 & 5.40 & 8.37 & 10.69 & 64.13 & 134.20 & 1.83 & 3.25 & 17.62 & 20.81 \\
        Q-7B-SFT & \colorbox{green!20}{2.65} & \colorbox{green!50}{4.47} & \colorbox{green!20}{7.39} & \colorbox{green!50}{9.08} & {59.15} & 148.47 & \colorbox{green!50}{1.50} &  \colorbox{green!50}{2.66} & \colorbox{green!20}{17.57} & \colorbox{green!20}{20.63} \\ \hline
        
        Q-3B-R1 & 3.21 & 4.93 & 11.49 & 13.75 & 67.82 & 127.66 & 1.70 & 3.08 & 19.75 & 22.43 \\
        Q-7B-R1 & 3.43 & 5.96 & 11.87 & 14.57 & 83.00 & 173.86 & 2.15 & 3.56 & 21.48 & 23.74 \\
        \hline
        Q-3B-FR & 3.06 & 5.00 & 8.26 & 10.42 & \colorbox{green!20}{52.51} & \colorbox{green!20}{114.42} & \colorbox{green!20}{1.60} & \colorbox{green!20}{2.91} &  \colorbox{green!50}{17.12} & 20.64  \\ 
        Q-7B-FR & \colorbox{green!50}{2.43} & \colorbox{green!20}{4.60} & \colorbox{green!50}{6.95} & \colorbox{green!20}{9.38} & \colorbox{green!50}{50.83} & \colorbox{green!50}{113.66} & 1.80 & 3.26 & 17.70 &  \colorbox{green!50}{20.53} \\
        \bottomrule
    \end{tabular}
    \end{threeparttable}
    \begin{tablenotes}
        \item \hspace{22mm} Q: Qwen-VL-2.5-Instruct; R1: model was trained with GRPO; FR: model was trained with FGRPR
    \end{tablenotes}
\end{table*}

\subsubsection{Models}
\begin{itemize}
\item \textbf{Baseline Models:} We evaluate several widely used benchmark models, including GPT-4o-2024-11-20 (\textit{GPT-4o}), Claude Sonnet-3.5-2024-06-20 (\textit{Sonnet-3.5}), LLaMA 2-70B (\textit{LLaMA-70B}), Intern-VL-26B (\textit{In-VL2-26B}), Qwen2.5-VL-3B (\textit{Q-3B}), and Qwen2.5-VL-7B (\textit{Q-7B}).
\item \textbf{SFT Models:} For comparative purposes, we perform supervised fine-tuning on Qwen2.5-VL-3B (\textit{Q-3B-SFT}) and Qwen2.5-VL-7B (\textit{Q-7B-SFT}).
\item \textbf{GRPO Models:} We conduct full-scale reinforcement learning training with the GRPO method on Qwen2.5-VL-3B (\textit{Q-3B-R1}) and Qwen2.5-VL-7B (\textit{Q-7B-R1}), employing the standard accuracy-based reward defined in equation \ref{equ: ra}.
\item \textbf{CrowdVLM-R1 Models (Ours):} We also perform full-scale RL training using our proposed FGRPR method on Qwen2.5-VL-3B (\textit{Q-3B-FR}) and Qwen2.5-VL-7B (\textit{Q-7B-FR}). Our fuzzy reward function is defined in equation \ref{equ: r}.
\end{itemize}

\subsubsection{Experimental Environment}

During training and testing for GRPO and FRGPR, we use 8 Nvidia L40S GPUs (each with 46GB of memory). The training dataset consists of 3803 images and 5 classes. The experiment is set to have a max prompt length of 1024, number of generations 8, per-device training batch size 6, gradient accumulation steps 2, data types bf16, and epochs 2. The testing consists of 600 images and 6 classes (one class is out of training samples). We use a batch size of 2 and data types bf16. It is worth noting that we use the same training and testing hyperparameters and prompt method across all the reinforcement learning models.

\subsection{Evaluation Metrics}

To evaluate the performance of the crowd counting task, we employ Mean Absolute Error (MAE) and Root Mean Squared Error (RMSE), which are standard metrics in the field. MAE measures the average absolute difference between predicted and actual crowd counts, providing an intuitive measure of accuracy:
\begin{equation}
    \text{MAE} = \frac{1}{n} \sum_{i=1}^{n} |y_i - \hat{y}_i|
    \label{equ:mae}
\end{equation}

where $y_i$ and $\hat{y}_i$ represent the ground truth and predicted counts, respectively, and $n$ is the total number of samples.

RMSE, defined as the square root of the mean squared differences, penalizes larger errors more heavily, making it particularly useful when significant deviations are critical for evaluation:

\begin{equation}
    \text{RMSE} = \sqrt{\frac{1}{n} \sum_{i=1}^{n} (y_i - \hat{y}_i)^2}
    \label{equ:rmse}
\end{equation}

While both metrics indicate model accuracy, MAE provides a straightforward interpretation of average error, whereas RMSE emphasizes larger errors by amplifying their impact. Lower values for both metrics signify better performance.

To comprehensively assess model performance, we employ additional evaluation metrics, including the coefficient of determination ($R^2$), Mean Absolute Percentage Error (MAPE), Pearson’s correlation coefficient, and Spearman’s rank correlation coefficient. Their definitions are provided below.

\subsubsection{Coefficient of Determination $R^2$}
The $R^2$ score measures how well the predicted values fit the actual values, indicating the proportion of variance explained by the model:

\begin{equation} R^2 = 1 - \frac{\sum_{i=1}^{n} (y_i - \hat{y_i})^2}{\sum{i=1}^{n} (y_i - \bar{y})^2} \end{equation}

where $y_i$ and $\hat{y_i}$ denote the actual and predicted values, respectively, $\bar{y}$ is the mean of the actual values, and $n$ is the total number of samples. An $R^2$ value closer to 1 indicates a better fit.

\subsubsection{Mean Absolute Percentage Error (MAPE)}
MAPE quantifies the relative prediction error as a percentage, providing an intuitive measure of accuracy:

\begin{equation} \text{MAPE} = \frac{1}{n} \sum_{i=1}^{n} \left| \frac{y_i - \hat{y}_i}{y_i} \right| \times 100 \end{equation}

A lower MAPE value indicates higher prediction accuracy. However, it is sensitive to small actual values, which can lead to inflated error percentages.

\subsubsection{Pearson’s Correlation Coefficient}
Pearson’s correlation coefficient ($r$) measures the linear relationship between predicted and actual values, ranging from -1 to 1:

\begin{equation} r = \frac{\sum_{i=1}^{n} (y_i - \bar{y}) (\hat{y_i} - \bar{\hat{y}})}{\sqrt{\sum{i=1}^{n} (y_i - \bar{y})^2} \sqrt{\sum_{i=1}^{n} (\hat{y}_i - \bar{\hat{y}})^2}} \end{equation}

A value close to 1 indicates a strong positive correlation, while a value close to -1 suggests a strong negative correlation.

\subsubsection{Spearman’s Rank Correlation Coefficient}

The Spearman's correlation ($\rho$) assesses the monotonic relationship between predicted and actual values by ranking the data points:

\begin{equation} \rho = 1 - \frac{6 \sum d_i^2}{n(n^2 - 1)} \end{equation}

where $d_i$ is the difference between the ranks of $y_i$ and $\hat{y_i}$, and $n$ is the number of observations. Unlike Pearson’s correlation, Spearman’s correlation captures non-linear relationships by evaluating rank similarity.


\begin{figure*}
    \centering
    \includegraphics[width=0.98\linewidth]{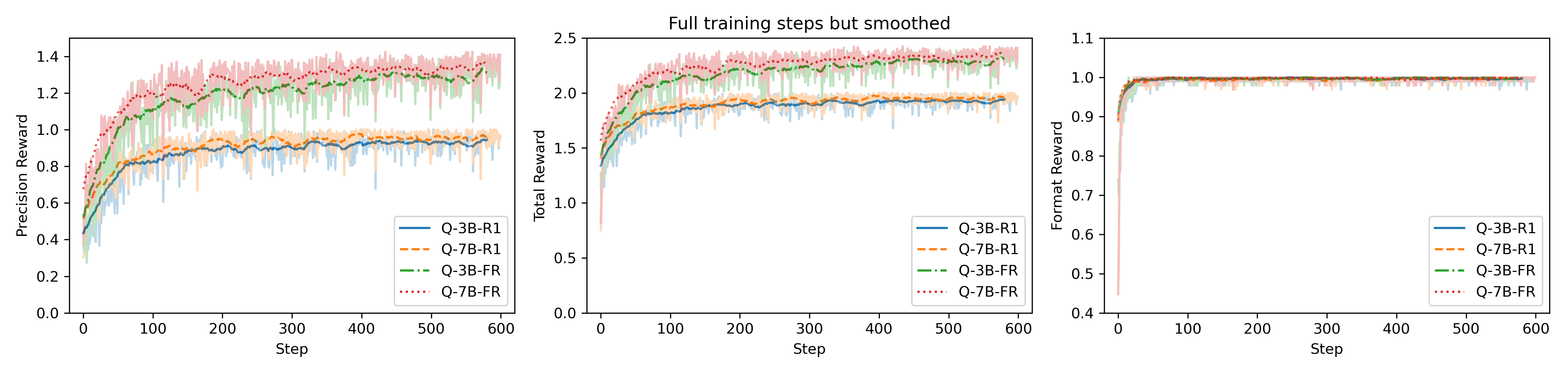}
    \caption{Training reward curves for different models show the same trend, while for precision reward, bigger models dominate the smaller models within the same method over the training steps after smoothed. And their format reward lines do not have any difference after 100 steps.}
    \label{fig:training_rewards}
\end{figure*}

\subsection{Results}
The overall model performance across different domains is shown in Table \ref{tab:accuracy}. \colorbox{green!50}{Green background}indicates the best results, while \colorbox{green!20}{light green}represents the second-best results. Among the baseline models, \textit{LLaMA-70B} demonstrates outstanding performance in four of the six domain datasets. Nevertheless, \textit{GPT-4o} and \textit{Sonnet-3.5} achieve lower MAE and RMSE scores compared to \textit{LLaMA-70B}. Specifically, \textit{GPT-4o} exhibits comparable performance to \textit{LLaMA-70B} on the out-of-domain dataset (\textit{Manatees}), displaying an MAE of approximately 5.7 and an RMSE around 10. On the other hand, \textit{Q-3B} and \textit{Q-7B} consistently show among the poorest metric results across all six datasets.

However, after traditional supervised fine-tuning, both \textit{Q-3B-SFT} and \textit{Q-7B-SFT} improve significantly. Specifically, they outperform not only their base versions but also all baseline models, including state-of-the-art models such as \textit{GPT-4o} and \textit{LLaMA-70B}, across five in-domain datasets. Although \textit{Q-7B-SFT} achieves a lower overall MAE compared to \textit{Q-3B-SFT}, its performance is not consistently better across these five in-domain datasets. Furthermore, \textit{Q-7B-SFT} presents a higher RMSE value, indicating that its prediction errors exhibit greater variance and instability.

After fine-tuning with GRPO methods employing the traditional $0/1$ reward function (rewarding predictions with $1$ if $\frac{\hat{y}-y}{y}<0.5$, otherwise $0$), the results indicate that although this approach exhibits some effectiveness, the models \textit{Q-3B-R1} and \textit{Q-7B-R1} show worse performance than their corresponding directly fine-tuned counterparts, \textit{Q-3B-SFT} and \textit{Q-7B-SFT}. This outcome reveals a limitation of applying such a GRPO-based reward approach specifically for the crowd counting task.

Models trained with our proposed FGRPR method, particularly \textit{Q-7B-FR}, exhibit state-of-the-art performance across all five in-domain datasets. They continue to reduce the MAE and RMSE compared to the previously best-trained model, \textit{Q-7B-SFT}, showing consistency in error reduction. For the 3B variant, \textit{Q-3B-FR} achieves better performance than the previous SFT model (\textit{Q-3B-SFT}) in two datasets, slightly worse performance in sheep counting, and equivalent performance in wheat head and traffic car counting datasets. Moreover, for the out-of-domain dataset, SFT and FGRPR show similar performance, which improved compared to their base models that show better performance than bigger models \textit{In-VL2-26B} and \textit{Sonnet-3.5}, but still can not achieve the best industry models like \textit{GPT-4o} and \textit{LLaMA-70B}. Considering the small size of our base models (only 3B and 7B), these results collectively prove the effectiveness of our proposed FGRPR reward function.


This effectiveness is further supported by Table \ref{tab:by_range}, where the models are evaluated based on different count ranges. In in-domain datasets, the SFT and FGRPR methods have similar performance with fewer objects, in ranges [1-20] and [21-50]. And all beat the most recent vision language and open-vocabulary counting method, T-Rex\cite{jiang2023t}, in out of domain dataset. However, as object counts exceed 50, FGRPR shows superior performance compared to SFT and GRPO, which is guided by the accuracy reward, equation \ref{equ: ra}. This highlights the effectiveness of our proposed reward function, equation \ref{equ: rp}, particularly the component ($\frac{\left|\hat{y} - y \right|}{y}$), which offers higher rewards for more accurate counts.

%% file: sec/limitation.tex
\subsection{Analysis}

\subsubsection{Model rewards}

In Fig. \ref{fig:training_rewards}, all four models demonstrate an upward trend in both precision and format rewards, as shown in the first-row subfigures where a simple moving average with a window size of 20 is applied across 600 training steps. The 7B models start at higher points than the 3B models, indicating superior initial performance, consistent with prior studies that larger models typically outperform smaller ones within the same family. And 3B model with FGRPR has similar start values to \textit{Q-7B-R1}.


All four models display a sharp increase within the initial 100 training steps, followed by fluctuating yet steady improvement in precision rewards over the subsequent 500 steps, reaching their peak around step 600. Notably, the FGRPR method demonstrates a sharper and steadier growth trajectory compared to the traditional GRPO method. Furthermore, the continuous upward trend shown by the FGRPR model at the training endpoint suggests the potential for further improvement with additional training data or a longer training period, whereas models trained via GRPO with traditional rewards plateau and exhibit little to no increase in reward values during the second half of training. Meanwhile, all four models have identical format reward curves, as they share the same format reward function.

Despite the clear overall trends revealed by smoothing, the shadows in Fig. \ref{fig:training_rewards} illustrate considerable fluctuations in reward values throughout training. At certain training steps, precision reward values drop abruptly before subsequently recovering and reconnecting with the general upward trend.

\subsubsection{Concentrated results}
Experiment results show that when the number of detected objects within the images is under 250, models prefer to predict some fixed count values.

\begin{figure}
    \centering
    \includegraphics[width=1\linewidth]{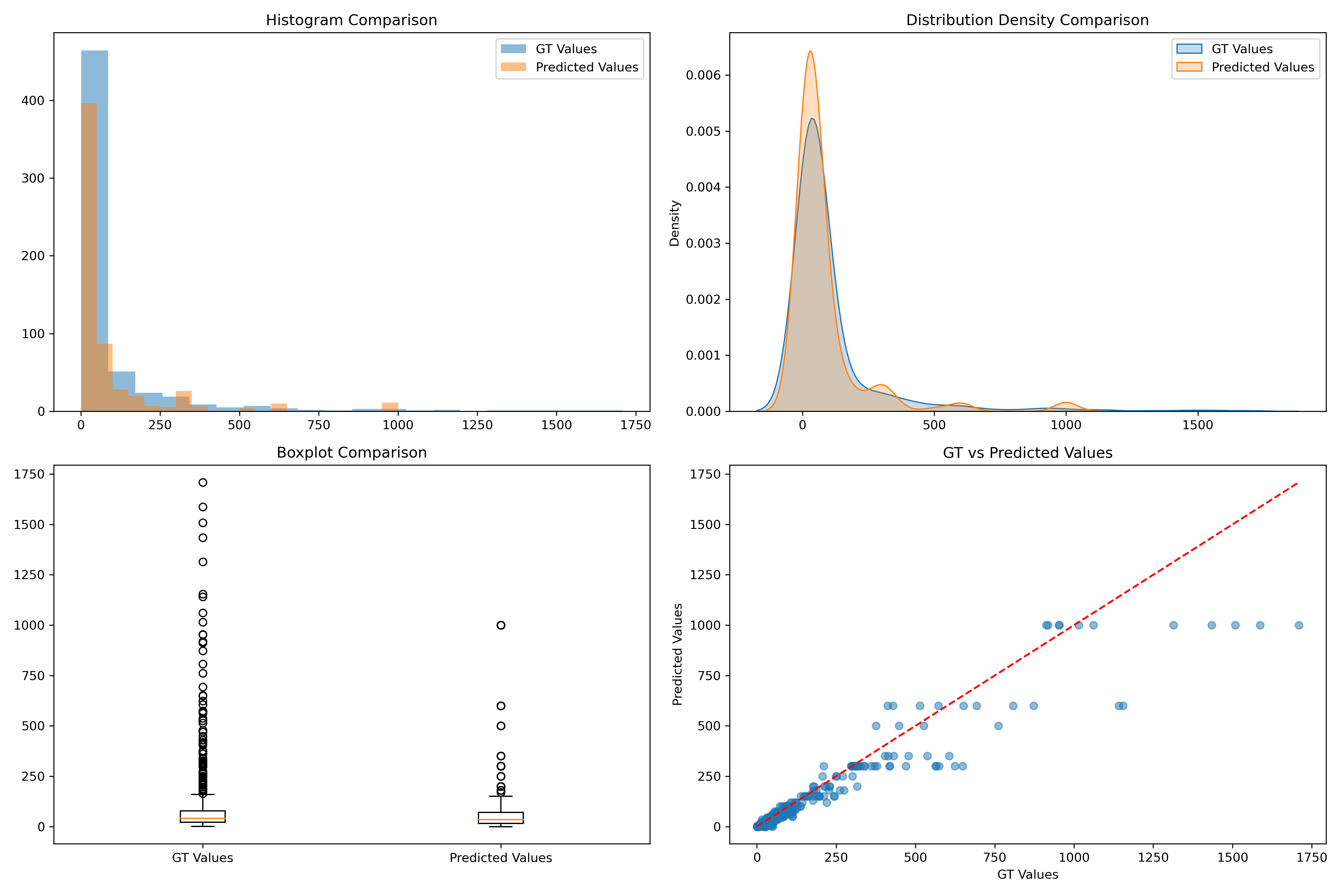}
    \caption{Distribution statistics comparison between true values and predicted values. Results from model \textit{Q-7B-RF}.}
    \label{fig:distribution_statistics}
\end{figure}

Fig. \ref{fig:distribution_statistics} presents distribution statistics for ground truth over prediction results. The histogram and density plots show that both distributions are highly right-skewed, with the majority of values concentrated below 250. However, the second row subfigures dramatically illustrate disparity between GT and predictions - GT values show numerous outliers spanning a much wider range, while predicted values display fewer outliers and compressed variability. 

\begin{table}
    \centering
    \caption{Additional prediction measurements}
    \renewcommand{\arraystretch}{1.5}
    \begin{tabular}{c c c c c }
        \hline
        Metrics &  $R^{2}$ & MAPE & Pearson & Spearman \\
        Value   &  0.8795 & 0.2389 & 0.9655 & 0.9497 \\
        \hline
    \end{tabular}
    \label{tab:additional_metrics}
\end{table}

Table \ref{tab:additional_metrics} reveals additional metrics for the prediction of the model. The $R^2$ value of 0.8795  indicates that the model explains nearly 88\% of the variance. Strong correlation coefficients (Pearson = 0.9655, Spearman = 0.9497) confirm that the model generally preserves rank ordering. And an MAPE of 23.89\% suggests predictions are, on average, off by about a quarter of the true value.

In conclusion, while the model shows a good performance in some ways, especially when below 250 counts, there are limitations. It produces concentrated predictions that do not reflect the true distribution's variability, and it also focuses heavily on certain values.

%% file: sec/conclusion.tex
\section{Conclusion}
The success of DeepSeek R1 highlights the effectiveness of RL in enhancing the reasoning abilities of LLMs for complex tasks, such as solving math problems. This approach has been extended to VLMs for tasks like interpreting mathematical figures. Previous work often used a simplistic "0/1" reward model, reducing answers to binary outcomes like "YES/NO". However, this reward function is inadequate for problems that require exact matches, such as crowd counting or position matching with intersection over union.

To address this, we proposed FGRPR, which leverages the robust RL method, GRPO, combined with a fuzzy reward system that assigns higher rewards for values closer to the ground truth. Our experiments demonstrate that the traditional "0/1" reward model with GRPO underperforms compared to SFT in these tasks. In contrast, FGRPR shows superior performance across all models, even surpassing \textit{GPT-4o} and \textit{LLaMA-70B} in in-domain datasets.

While FGRPR performs comparably to SFT in out-of-domain datasets with smaller target values, it excels with larger targets. This makes sense since traditional RL methods disregard the distinction between $\frac{y}{2}$ and $\frac{3y}{2}$. When $y$ is smaller, although our method remains competitive, it does not always provide an advantage—an aspect that could be improved in the future.

Although this research specifically evaluates the effectiveness of FGRPR within the domain of crowd counting, the framework we propose can easily be generalized and applied to other estimation tasks where fuzzy reward evaluations are beneficial.